\documentclass[journal]{IEEEtran}
\usepackage{amsmath,amsfonts}
\usepackage{algorithmic}
\usepackage{algorithm}
\usepackage{array}
\usepackage[caption=false,font=normalsize,labelfont=sf,textfont=sf]{subfig}
\usepackage{textcomp}
\usepackage{stfloats}
\usepackage{url}
\usepackage{verbatim}
\usepackage{color, soul}
\usepackage[dvipsnames]{xcolor}
\usepackage{graphicx}
\usepackage{cite}
\usepackage[symbol]{footmisc}
\usepackage{booktabs}

\usepackage{multirow}
\newcommand{\revAdd}[1]{#1}

\newcommand{\revAddd}[1]{#1}

\begin{document}

\title{BERTO: Intent-Driven Network Time Series Forecasting via Natural Language Operator Preferences}

\author{Nitin Priyadarshini Shankar, Vaibhav Singh, Sheetal Kalyani, and Christian Maciocco
\thanks{Nitin P. Shankar is with Intel Labs, Bengaluru, India, and the Department of Electrical Engineering, Indian Institute of Technology Madras, Chennai, India. This work was conducted while Nitin P. Shankar was at Intel Labs. (e-mail: nitin.priyadarshini.shankar@intel.com, ee20d425@smail.iitm.ac.in)}
\thanks{Vaibhav Singh is with Intel Labs, Bengaluru, India. (e-mail: vaibhav1.singh@intel.com)}
\thanks{Sheetal Kalyani is with the Department of Electrical Engineering, Indian Institute of Technology Madras, Chennai, India. (e-mail: skalyani@ee.iitm.ac.in)}
\thanks{Christian Maciocco is with Intel Labs, Oregon, United States. (e-mail: christian.maciocco@intel.com)}}
\maketitle

\begin{abstract}


\revAdd{Traditional cellular traffic forecasting models are optimized for minimizing symmetric errors, leaving them indifferent to shifting operational priorities. To bridge this gap, we introduce BERTO, a BERT-based framework for traffic prediction and energy optimization in cellular networks. Built on transformer architectures, BERTO achieves high prediction accuracy while enabling a single fine-tuned model to operate across multiple forecasting regimes via natural-language operator prompts. By combining a Balancing Loss Function (BLF) with prompt-based conditioning, BERTO adaptively shifts its forecasting bias toward underprediction or overprediction depending on the operator’s desired trade-off between power savings and service quality. This allows the same model to dynamically generate different decision-aware forecasts without retraining or modifying model parameters. Experiments on real-world datasets demonstrate that BERTO can operate across a flexible range of approximately \(1.4\) kW in power consumption while balancing \(9\times\) variation in service level agreement (SLA) violations, making it well suited for intelligent RAN deployments.}

\end{abstract}

\section{Introduction}

Time series data is ubiquitous across all layers of modern communication networks. From Radio Access Network (RAN) KPIs (Key Performance Indicators) to core metrics such as CPU utilization and user traffic patterns, this data forms the foundation for understanding and optimizing network behavior. \revAdd{Traditional forecasting models are excellent at predicting what will happen, but they are frequently ``intent-blind," failing to understand why the prediction matters to the operator. This creates a dangerous decoupling between the forecasts and real-world utility. For instance, a model predicting a $20\%$ spike in traffic is technically accurate, but its value changes entirely depending on the operator's intent: under a ``User Experience First" mandate, this forecast should trigger immediate resource provisioning; under an ``Energy Saving" goal, it might instead trigger a graceful throttling of non-essential services. By transitioning to operator-intent-based forecasting, we ensure predictions-based decisions are not just accurate but purposeful.
}

Traditionally, time series forecasting has relied on statistical models, such as ARIMA \cite{1203886} and Exponential Smoothing (ETS) \cite{4446113}, as well as deep learning techniques, including RNNs and LSTMs \cite{10.1145/3291533.3291540}. These models often struggle with generalization, long-range dependencies, and integrating contextual information. Transformer-based models have emerged as strong alternatives due to their self-attention mechanism, which effectively captures long-range dependencies. In \cite{zhou2021informer}, the Informer employs ProbSparse self-attention and distilling to reduce the quadratic complexity of Transformers. Autoformer \cite{NEURIPS2021_bcc0d400} enhances \cite{zhou2021informer} with series decomposition and Auto-Correlation, replacing self-attention for efficient series modeling. Autoformer \cite{10615438} forecasts wireless network traffic at short time intervals and uses the forecast to dynamically orchestrate and deploy either an RL-driven traffic steering xApp or a cell-sleeping rApp.  \cite{10114636} proposes GLSTTN, which combines transformer modules and densely connected CNNs to achieve state-of-the-art accuracy in city-level cellular traffic prediction across multiple traffic types. \revAdd{However, even these state-of-the-art architectures share a common limitation: they are optimized exclusively for minimizing mathematical error. By treating all forecasting errors as equally "bad," they fail to account for the asymmetric risks inherent in network operations, such as the high cost of a service outage versus the minor cost of temporary over-provisioning. In this work, we bridge this gap by proposing a model that goes beyond objective error metrics and dynamically adjusts its predictions to align with operator-defined preferences expressed in natural language.}



With the emergence of Large Language Models (LLMs) built on transformer architectures, a new paradigm has emerged in time series forecasting. These models not only support powerful temporal modeling but also offer capabilities such as zero-shot and few-shot learning, and can be finetuned for specific domains. \revAdd{Recent research has begun to exploit these strengths:} \cite{10901784} analyzes temporal traffic patterns to select key historical timespans and clusters similar cells, then encodes their multi-timespan data into LLM prompts for prediction. \cite{11206450} presents SpectrumLLM, the first architecture to align radio-spectrum state time series with LLM text space via tokenization and embedding. \revAdd{However, the most profound advantage of LLMs lies in their instruction-following capability to condition numerical forecasts on linguistic operator intent. This allows the model to act as a bridge between high-level operational goals and low-level network data. For instance, by providing a natural language prompt that prioritizes ``Energy Efficiency", the operator can guide the model to generate a forecast that emphasizes potential low-traffic windows, facilitating more aggressive cell-offloading or deep-sleep cycles. In this work, we move beyond simple data encoding and address a fundamental question: Can we leverage LLMs' linguistic adaptability to transform a network forecaster into a goal-aware system that predicts the future through the lens of specific operator intent?} 



\revAdd{To address this need, we introduce BERTO (BERT-based Operator-intent forecaster), a transformer-based architecture fine-tuned specifically for the nuances of cellular network traffic. Unlike static models, BERTO is designed to consume operator objectives in natural language as an integral part of the input prompt. By leveraging a Balancing Loss Function (BLF) [11], a single instance of BERTO can adaptively shift its predictive bias in real-time, optimizing for ``Power Savings" or ``Service Quality" as the situation demands. This allows for a flexible, goal-oriented forecasting mechanism that bridges the gap between linguistic intent and numerical precision.}

\revAdd{The main contributions of this work are the following.
\begin{enumerate}
    \item Intent-Aware Architecture: We propose a BERT-based framework that integrates natural-language prompts with time-series data, enabling the first ``goal-aware" forecaster for cellular networks.
    \item Dynamic Adaptability via BLF: We demonstrate how a single fine-tuned model can be made responsive to conflicting operator preferences using a balancing loss mechanism, eliminating the need for multiple task-specific models.
    \item Extensive Benchmarking: We provide a rigorous performance evaluation against a diverse suite of baselines, including classical models (ARIMA), deep learning (LSTM, FNN), and state-of-the-art transformer architectures (Autoformer, Chronos).
\end{enumerate}}


\section{Problem Formulation}


Consider a cluster of co-located cells within a shared service area $A$, where each cell $i$ serves a set of users and has a specific load at a given time. {Co-located cells refer to multiple cellular base stations positioned at the same physical site, typically sharing infrastructure and covering overlapping areas.} Forecasting a cell's future load enables the network to adapt, enhancing performance and efficiency. \revAdd{Let $\mathcal{I}$ denote the natural language string representing the active operator intent. The time series prediction problem for the $i^{\text{th}}$ cell is to compute intent conditioned forecast $x_{t}$ given $x_{t-h:t-1}$ along with other statistics such as $\mu_{x_{t}}$ and $\sigma_{x_{t}}$ calculated based on past load information}. In this work, we focus on short-term prediction, so $x_{t}$ is a scalar. If the predicted value is $\hat{x_{t}}$, the prediction model is $f_\theta(.)$ where $\theta$ are the parameters of the model and the loss function used to train the model is $\mathcal{L}(.)$. The optimization problem can be defined as 

\begin{equation}
    \hat{\theta} = \text{argmin}_\theta(\mathcal{L}(x_{t}, \hat{x_{t}}))    
\end{equation}

Where \revAdd{$\hat{x_{t}} = f_\theta(x_{t-h:t-1},\mathcal{I}, \mu_{x_{t}}, \sigma_{x_{t}})$ and $\hat{\theta}$ is the optimized model parameter after training. After model training, the intent-aware predicted value is used to determine the optimal sleep or active state configuration for each cell, thereby minimizing power consumption while maintaining service quality. Taking into account the intent in the forecasted value ensures that the generated prediction is already pre-biased to match the operator's downstream operational constraints}. For a cluster with $|B|$ co-located cells and a maximum of $K$ sleep states per cell, the state space for the cluster can be as large as $K^{|B|}$. This makes the problem computationally intensive in dense network scenarios where $|B|$ is large. To ensure coverage and performance, constraints are applied to sleep-state values, such as prioritizing cells with higher center frequencies or licensed spectrum in scenarios with reduced traffic. The priority order of cells, denoted as $\mathcal{C}$, is assumed to be known or learned over time. The sleep state configuration for the cluster, represented by the vector $\mathbf{z}$, must satisfy the constraints $\mathbf{z} \in \mathcal{C}$. \revAdd{The predicted cell load vector $\mathbf{\hat{x}}$ is used to minimize power consumption $P(\mathbf{x}, \mathbf{z})$ while maximizing service quality $f(\mathbf{x}, \mathbf{z})$. The optimization problem is:}
\begin{equation}
    \max_{\mathbf{z} \in \mathcal{C}} f(\mathbf{x}, \mathbf{z}) - P(\mathbf{x}, \mathbf{z}),    
\end{equation}
where the terms $f(\mathbf{x}, \mathbf{z})$ and $P(\mathbf{x}, \mathbf{z})$ capture the trade-off between service quality and power savings. The major challenge in solving the problem (1) is to obtain an accurate estimate of $\mathbf{x}$. The system model relies on a rich Radio Access Network (RAN) data pipeline, including historical load metrics such as the number of connected users and resource block usage. Performance metrics, such as cell throughput and service quality outages, as well as metadata, including holidays and special events, are also leveraged. In summary, the future load is predicted using current RAN performance data, enabling actions that conserve power and ensure service quality. 



\begin{figure}
    \centering
    \includegraphics[scale=0.69]{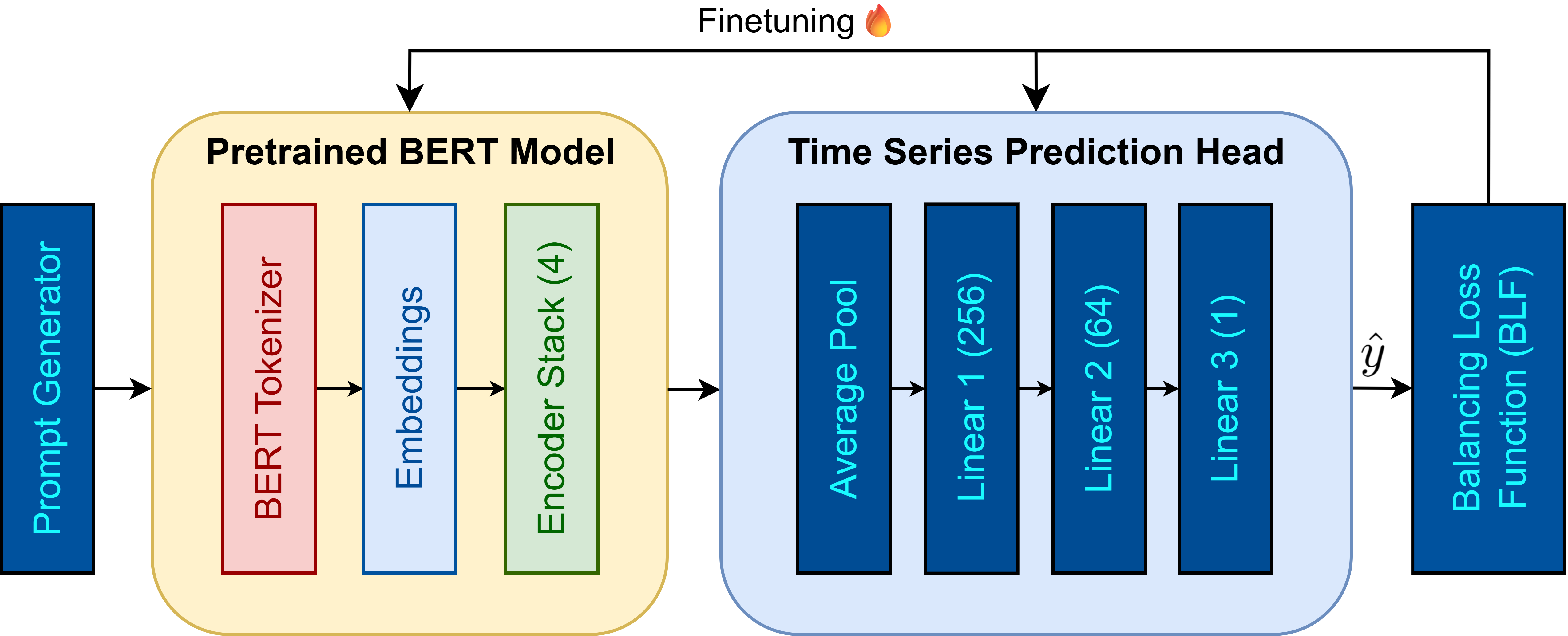}
    \caption{BERTO Architecture.\label{fig:bertts}} 
\end{figure}

\section{Proposed Method - BERTO}

The proposed architecture comprises of a \revAddd{transformer-based} pre-trained BERT \cite{devlin2018bert} model, which is fine-tuned with time series prompts to predict future loads using a Time Series Prediction (TSP) head. For this work, we utilize a BERT model with $4$ encoder layers and a hidden layer dimension of $256$. \revAdd{To adapt BERT for time series forecasting, we introduce two task-specific components: the Prompt Generator and the Time Series Prediction Head. Finally, to make BERTO intent-aware, we have used the BLF during finetuning.} 

\subsection{Prompt generation}

\begin{figure}
    \centering
    \includegraphics[scale=0.12]{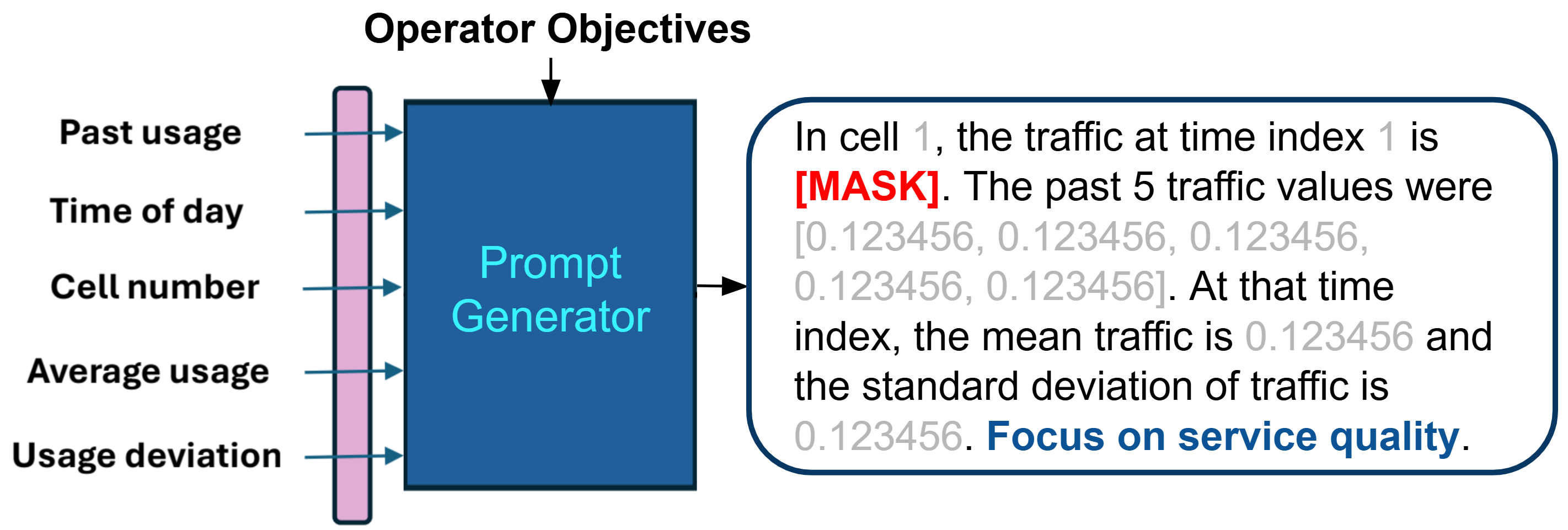}
    \caption{Prompt generation.\label{fig:prompt}} 
\end{figure}

\revAddd{We propose to leverage} essential factors pertinent to the prediction task, such as \revAdd{ \textit{Natural Language Operator Objective}}, \textit{Past Usage}, \textit{Time of Day}, \textit{Cell Number}, \textit{Average Usage}, and \textit{Usage Deviation}, as illustrated in Fig. \ref{fig:prompt}. These features are processed by the \textbf{Prompt Generator}, which formulates a structured textual representation to construct the prompt. In this representation, the current traffic value is obscured using the (\texttt{[MASK]}) token, designating it as the target label to be predicted. Consequently, the generated prompt serves as the input to the BERT model, functioning similarly to feature inputs in traditional machine learning algorithms.

\subsection{Time Series Prediction head}

A custom regression head was added to adapt the pre-trained BERT model for time series analysis. The prompts created by the prompt generator are then passed on to the BERT tokenizer, which converts them into token IDs and attention masks, as shown in Fig. \ref{fig:bertts}. Then, the token IDs and attention masks are processed through the base BERT model to extract contextual embeddings from their final hidden states. \revAddd{We propose using a} 2D average pooling operation with a kernel size of \(3 \times 3\) and a stride of $3$ to reduce dimensionality and capture key features \revAddd{at the output of the BERT model}. The pooled features are passed through a series of fully connected layers. The first linear layer maps the pooled features to a dimension of $512$. A subsequent linear layer reduces the dimensionality to $64$, followed by a final linear layer that outputs a single value representing the regression prediction for the time series task. These linear layers are designed to extract temporal patterns without significantly increasing complexity. \revAddd{Finally,} the predicted value is fed into the BLF loss function to compute the loss, which is then used for backpropagation to fine-tune both the pretrained BERT model and the TSP head together.

\subsection{Balancing Loss Function for BERTO}

The BLF \cite{9930603} is an asymmetrical loss function designed to balance underprediction and overprediction in machine learning models. It introduces a tunable parameter, \(q\), that controls the penalization asymmetry. Mathematically, BLF is defined as:

\begin{equation}
    \text{BLF} = \max\left\{  \frac{q \cdot (y - \hat{y}) }{q+1}, \frac{(\hat{y} - y)}{q+1} \right\},    
\end{equation}

where \(y - \hat{y}\) represents the prediction error, with \(y\) being the true value and \(\hat{y}\) the predicted value. The parameter \(q\) determines the relative weight of underprediction versus overprediction. For \(q > 1\), the loss increases significantly for underpredictions \revAddd{$(y - \hat{y} > 0)$}, discouraging such errors. Conversely, for \(q < 1\), overpredictions \revAddd{$(y - \hat{y} < 0)$} are penalized more heavily. This flexibility enables BLF to tailor the model's behavior to meet the specific needs of each application. \revAddd{Asymmetric loss functions \revAddd{\cite{8328873}} are crucial in practical applications since the cost of overprediction or underprediction is not the same. In contrast, the MSE loss implicitly assumes that both are equivalent.} \revAdd{An illustrative example of how BLF works can be found in Appendix D (see Supplementary Material).}

\revAdd{Replacing MSE with BLF and finetuning the model with different values of \(q\) enables BERTO to bias its predictions toward either underprediction or overprediction, depending on the operator objective. BERTO does not assume that future traffic demand changes with operator preference; instead, it learns forecasting biases that better align predictions with the desired trade-off between power savings and service quality. This behavior is controlled by natural-language operator prompts appended to the input, as shown in Fig.~\ref{fig:prompt}. During finetuning, the same traffic samples are paired with different prompts (with operator objectives) and corresponding BLF parameters \(q \in \{0.1,0.5,1,5,10\}\), allowing a single BERTO model to learn multiple operating modes. More details of the Training Strategy can be found in Appendix B (see Supplementary Material). Consequently, the model can dynamically shift across different power-saving and service-quality trade-offs at inference time without retraining or modifying its parameters, as illustrated in TABLE~\ref{tab:bertoo}.}


\subsection{Cell On-Off scheme}
 
The cell on-off scheme is a load management strategy \revAddd{\cite{10570553}} that reduces energy consumption by selectively switching off one of the cells in an underutilized co-located cell pair. This scheme is active if the estimated loads of both cells are less than a predefined threshold. Let $\hat{L_{1}}$ and $\hat{L_{2}}$ be the estimated loads of cells $1$ and $2$. If $(\hat{L_{1}} + e.\hat{L_{2}}) \leq L_{\text{th}}$, cell $2$ is switched off to save power. Here, $L_{\text{th}}$ is the load threshold, and \(e\) is the spectral efficiency ratio of the cell pair. The remaining active cell absorbs the load of the turned-off cell, ensuring uninterrupted service to users. This scheme relies on accurate load predictions and a well-defined threshold to determine optimal switching points, thereby minimizing energy consumption while maintaining quality of service.

\section{Experiments}

\subsection{Dataset}

\revAddd{Milan’s open-source Telecom Italia dataset provides anonymized call detail records (CDRs) that capture the city’s spatial and temporal communication dynamics. It contains aggregated call, SMS, and internet activity data, organized by geographic grid and time interval. The city is divided into $10{,}000$ grids of roughly $200 \ \text{m}$ per side, with CDR statistics estimated per grid. In this work, we use internet activity for traffic prediction, as it represents the dominant component of throughput. The dataset has a $10$-minute granularity (144 samples per day) spanning two months. As shown in \cite{9930603}, future load is highly correlated with recent past samples and with neighbouring grids. Combined with the small grid size, this justifies mapping each grid to a single cell.} Additionally, due to the high correlation, we can assume that adjacent cells are co-located, where $i$ corresponds to the low-frequency cell and $i+1$ corresponds to the high-frequency cell, with $i \in \{0, 2, 4, \dots \} $. The high-frequency cell is turned off to conserve resources according to the cell on-off scheme discussed in the next section. \revAddd{This co-location assumption is necessary because cell on–off decisions rely on the cell pair covering the same area so that one cell can be safely switched off while the other maintains service.}




\begin{table}
    \caption{Performance comparison}
    \centering
    \begin{center}        
    \begin{tabular}{|p{4.25cm}|p{0.75cm}|p{2.5cm}|}
    \hline
    \textbf{Model} & \textbf{MSE} & \textbf{Inference Time (ms)} \\ 
     \hline
    Previous Value Predictor  & $10.78$ & 0 \\
    FNN \cite{329294} (200 Parameters) & $8.09$ & $0.078$ \\
    LSTM \cite{10.1145/3291533.3291540} (750 Parameters) & $7.98$ & $0.266$\\
    ARIMA \cite{1203886} (p = 1, d = 0, q = 3)& $8.59$ & $3.75$ \\
    Autoformer \cite{NEURIPS2021_bcc0d400} (13M Parameters) & $8.95$ & $8.43$\\
    Chronos \cite{ansari2024chronos} (20M Parameters) & $8.61$ & $31.76$\\
    \textbf{{BERT$_{\text{MSE}}$}} (12M Parameters) (Ours) & $\mathbf{{7.65}}$ & $4.87$\\
    \hline
    \end{tabular}
    \label{tab:mse}
    \end{center}
\end{table}

\subsection{Experimental setup}

The simulation was carried out on a system with $2$ Intel Xeon processors and NVIDIA RTX 3090 GPUs. Training was conducted on data spanning $11$ days, while testing was performed on data from the subsequent $3$ days. The experiments were conducted on $428$ cells from the Telecom Italia Dataset. \revAddd{This work limits the cluster size to $2$ cells \cite{9930603}, to simplify the implementation and focus mainly on load prediction. However, the proposed solution can be extended to larger clusters.} \revAdd{More details on the simulation parameters are present in Appendix C (see Supplementary Material)\footnote{All appendices referenced in the main text are provided in the
Supplementary material}.}

\subsection{Baseline Methods}

    \revAdd{ To establish a minimum performance benchmark, we employ a previous value predictor, a naive heuristic that assumes the next network state is identical to the current one. To capture the periodic nature of network traffic, \cite{1203886} utilizes seasonal ARIMA models with multiple periodicities for effective forecasting of wireless traffic streams. In addition, standard deep learning baselines, including Long Short-Term Memory (LSTM) networks \cite{10.1145/3291533.3291540} and Feedforward Neural Networks (FNNs) \cite{329294}, are considered for comparison. Transformer-based predictors, such as Autoformer \cite{NEURIPS2021_bcc0d400} and Chronos \cite{ansari2024chronos}, further extend this set of baselines: Autoformer introduces a decomposition-based architecture with a periodicity-driven Auto-Correlation mechanism to model sub-series dependencies. At the same time, Chronos leverages a pretrained transformer framework with in-context learning to enable zero-shot forecasting across univariate and multivariate time series.}

\begin{table}
    \caption{BERTO comparison}
    \centering
    \begin{center}        
    \begin{tabular}{|p{1.05cm}|p{2.3cm}|p{0.27cm}|p{1.6cm}|p{1.5cm}|}
    \hline
     \textbf{Model} & \textbf{Operator Prompts} & $\mathbf{q}$ & \textbf{Total Power Savings (W)} & \textbf{Average Throughput loss (\%)}\\
     \hline
       {BERT$_{\text{MSE}}$} & - & - &$0$ (Baseline)$^{\ddagger}$ & $0.057$\\
       \hline
           {LSTM$_{\text{MSE}}$} & - & - &$13.13^{}$ & $0.062$\\
        \hline
    {FNN} & - & - &$-460.95^{\ddagger}$ & $0.070$\\
        \hline
        & ``Focus highly on power savings" & 0.1 & $\mathbf{\textcolor{Green}{714.76}}$& ${0.252}$ \\ \cline{2-5}
        & ``Focus on power savings" & 0.5& $219.04$& $0.099$\\ \cline{2-5}
        BERTO & ``No specific focus"& 1 & $115.56$& $0.066$\\
         \cline{2-5}
         & ``Focus on service quality"& 5 &$-410.88^{\ddagger}$ & $0.041$\\ \cline{2-5}
         & ``Focus highly on service quality"& 10 &${-731.88}^{\ddagger}$ & $\mathbf{\textcolor{Green}{0.026}}$ \\ 
         \hline
    \end{tabular}
    \label{tab:bertoo}
    \end{center}
\end{table}

\subsection{Performance Metrics}

\subsubsection{Power model and savings calculation} The power consumption model evaluates the energy usage of cellular networks in both active and inactive states. For co-located cell pairs, the power consumption for the active state is extrapolated from \cite{7145603} as: $P_{\text{on}} = 2A + B \ (L_1 + L_2) \text{  W},$

where \(L_1\) and \(L_2\) represent the loads of the cells. When one cell is turned off, the power consumption is reduced to: $P_{\text{off}} = A + B \ (L_1 + e. L_2) \text{  W},$

 The power savings are computed as the difference between the active and inactive power states, providing insights into the cell's on-off scheme energy efficiency. 

\subsubsection{Throughput loss calculation}
The throughput loss model is defined as:
\begin{equation}
R_{\text{loss}} = 
\begin{cases} 
0 & \text{if } ({L_{1}} + e.{L_{2}}) \leq L_{\text{max}}, \\
L_1 + e.L_2 - L_{\text{max}} & \text{otherwise}.
\end{cases}    
\end{equation}

The throughput loss ($R_{\text{loss}}$) is calculated when one cell is turned off and when the total load of the cell pair is greater than the maximum load serviceable by a single cell ($L_{\text{max}}$). 

\subsection{Simulation Results}



TABLE \ref{tab:mse} presents the MSE performance comparison of our proposed model, {BERT$_{\text{MSE}}$}, with baseline and SOTA methods. {BERT$_{\text{MSE}}$} achieves the lowest MSE of $7.65$, demonstrating a $29.1\%$ improvement over the Previous Value Predictor (MSE: $10.78$) and a $4.13\%$ improvement over the next best-performing method, LSTM (MSE: $7.98$). Additionally, {BERT$_{\text{MSE}}$} outperforms methods like Chronos \cite{ansari2024chronos} and Autoformer \cite{NEURIPS2021_bcc0d400}, which have significantly higher parameter counts, highlighting the efficiency of our approach.

TABLE \ref{tab:bertoo} showcases the flexibility of BERTO, which extends {BERT$_{\text{MSE}}$} by incorporating operator prompts for adaptive predictions. \revAddd{By adjusting the BLF parameter $q$ based on operator prompts, the model selectively emphasizes underprediction or overprediction, aligning its behaviour with the operator’s specific objective.} BERTO enables a single model to perform a wide range of predictions, balancing power savings and throughput loss. For instance, BERTO achieves a maximum power savings of $714.76$ W with the prompt ``Focus highly on power savings", corresponding to an average throughput loss of $0.252\%$. In contrast, with the ``Focus highly on service quality" prompt, the model minimizes throughput loss to $0.026\%$ while sacrificing power savings.

This flexibility is critical for adapting to dynamic operator objectives, as a single BERTO model is capable of operating across a power range of $-731.88$ W\footnote[3]{Here, a negative value indicates an increase in power consumption relative to the baseline measured across 428 cells.} to $714.76$ W and a throughput loss range of $0.026\%$ to $0.252\%$. These results show that BERTO can operate over a range of $1446.64$ W with a $9.69 \times$ flexibility in average throughput loss for $428$ cells. These results underline the adaptability and robustness of BERTO, making it a highly versatile solution for balancing conflicting objectives in real-world scenarios without requiring the operator to tune any parameters.

\revAdd{Fig.~\ref{fig:pareto} further illustrates this adaptability by showing the Pareto front obtained when evaluating the same testing dataset with different operator prompts appended to the input. Note that the $q$ values are unused during testing. Each prompt shifts the operating point of the same fine-tuned BERTO model toward a different trade-off region between power savings and throughput loss. This demonstrates that BERTO supports multiple operating regimes without retraining or modifying model parameters, enabling prompt-driven adaptation to changing network objectives.}


{In addition to prediction accuracy, we evaluated the inference time (for one sample) of all models to assess their suitability for real-time network deployment. Among all methods, BERTO demonstrates an effective balance between model complexity and latency, achieving an inference time of $4.87\ \mathrm{ms}$. Simpler models, such as FNN and LSTM, exhibit lower latency but with reduced representational capacity. Autoformer, which includes a decoder as part of its architecture, records a moderate inference time of $8.43\ \mathrm{ms}$. In contrast, Chronos incurs the highest latency of $31.76\ \mathrm{ms}$ due to its larger architecture and probabilistic forecasting design. }


\begin{figure}
    \centering
    \includegraphics[scale=0.32]{fig/pareto_plot.png}
    \caption{Pareto front of total power savings versus average throughput
    loss for BERTO swept over the BLF shape parameter $q \in \{0.1, 0.5, 1, 5, 10\}$ vs MSE baselines. \label{fig:pareto}} 
\end{figure}

\section{Conclusion}


The proposed BERT-based frameworks excel in network time series prediction. {BERT$_{\text{MSE}}$} delivers precise predictions with a low MSE, while BERTO introduces flexibility by balancing power savings and throughput loss through operator prompts. These results highlight BERTO's potential as a dynamic and adaptable solution for energy management in cellular networks. {Also, BERTO enables an operator to modify the model's prediction solely based on the input prompts without needing to modify its parameters.} Future work could focus on scaling the model using various methods to reduce its complexity and integrating additional real-time factors to enhance its practical applicability and flexibility of operation.

\bibliographystyle{IEEEtran}
\bibliography{refs}

\setcounter{section}{0}
\renewcommand{\thesection}{\Alph{section}}
\renewcommand{\thesubsection}{\thesection.\arabic{subsection}} 

\clearpage

\maketitle
\section*{Appendix A}
\subsection*{Transformer Basics}

 A Transformer consists of an encoder-decoder architecture. The encoder processes the input sequence to generate contextualized embeddings, while the decoder generates the output sequence conditioned on them. Each encoder and decoder block comprises two primary components: multi-head self-attention and position-wise feed-forward networks. Layer normalization and residual connections are used to stabilize training. The self-attention mechanism computes the relationships between tokens using three learnable matrices: Query ($\mathbf{Q}$), Key ($\mathbf{K}$), and Value ($\mathbf{V}$). The output is calculated as:
\begin{equation}
\text{Attention}(\mathbf{Q}, \mathbf{K}, \mathbf{V}) = \text{softmax}\left(\frac{\mathbf{Q}\mathbf{K}^T}{\sqrt{d_k}}\right)\mathbf{V},
\end{equation}
Where $d_k$ is the dimensionality of the key vectors, positional encoding is added to the input embeddings to retain sequential information. The Transformer architecture's scalability and parallelizability have made it the foundation for state-of-the-art models, such as BERT and GPT.


\subsection*{BERT}


\revAddd{The Bidirectional Encoder Representations from Transformers (BERT) model [10] is a state-of-the-art language representation model for natural language processing. Unlike traditional unidirectional models, BERT uses a bidirectional Transformer architecture with self-attention, allowing it to capture contextual relationships between words. It is pretrained on two tasks: Masked Language Modeling (MLM), where randomly masked tokens are predicted, and Next Sentence Prediction (NSP), which determines whether two sentences are sequentially related. These tasks enable BERT to learn deep contextual embeddings, making it effective for downstream applications such as question answering and sentiment analysis. MLM and NSP heads are attached to the encoder output to perform their respective objectives. In this work, we introduce a Time Series Prediction (TSP) head, jointly fine-tuned with BERT, for network time series prediction.}

\section*{Appendix B}
\subsection*{Training Strategy}
\label{subsec:training_strategy}

A key design principle of BERTO is that the underlying traffic
generation process is independent of operator preference. The
historical traffic volume in the dataset is a deterministic physical
measurement and is not modified, reweighted, or reinterpreted across
operating points. The same traffic samples are used for all settings.
What changes during training is the \emph{objective function}, which
encodes the asymmetric cost of forecasting errors in cellular
operation.

Under standard Mean Squared Error (MSE), overprediction and
underprediction are penalized equally, and the optimal predictor is
centered around the conditional mean. However, in a practical network
control, the cost of these two errors is asymmetric: underprediction
may trigger overly aggressive cell shutdown and degrade service
quality, while overprediction may keep additional cells active and
increase energy consumption. The most useful forecast for downstream
control may therefore deviate from the conditional mean. BERTO
captures this asymmetry through the proposed Balancing Loss Function
(BLF), parameterized by~$q$:
\begin{itemize}
    \item $q>1$: penalizes underprediction more strongly, biasing
    forecasts toward service-quality-oriented operation;
    \item $q=1$: recovers an approximately symmetric penalty;
    \item $q<1$: penalizes overprediction more strongly, biasing
    forecasts toward power-saving-oriented operation.
\end{itemize}

To make BERTO adaptive to operator intent, during fine-tuning, the
model is exposed to the \emph{same time-series samples multiple
times}, each paired with a different natural language prompt and the
corresponding $q$ value in the loss function. Representative pairings
used in our training pipeline are summarized in
TABLE~\ref{tab:prompt_q_pairs}.

\begin{table}[t]
\centering
\caption{Representative operator prompts and corresponding BLF
parameter $q$ used during fine-tuning.}
\label{tab:prompt_q_pairs}
\begin{tabular}{ll}
\toprule
\textbf{Operator Prompt} & \textbf{$q$} \\
\midrule
``Focus highly on service quality'' & $10$ \\
``Focus on service quality'' & $5$ \\
``No specific focus''               & $1$  \\
``Focus on power savings''   & $0.5$ \\
``Focus highly on power savings''   & $0.1$ \\
\bottomrule
\end{tabular}
\end{table}

This procedure teaches the model to associate operator intent,
expressed in natural language, with the appropriate forecasting bias.
As a result, a \emph{single} BERTO model generalizes across multiple
operating objectives, avoiding the alternative of training, storing,
and switching between multiple specialized predictors (e.g., one LSTM
per preference). This design is practical for modern deployments,
since operators typically have sufficient compute at the base station
controller, edge cloud, or base station cluster to support
millisecond-level centralized inference.

\section*{Appendix C}
\subsection*{Simulation Parameters}

\begin{table}[!ht]
    \caption{Simulation Parameters for BERTO}
    \centering
    \begin{center}        
    \begin{tabular}{|p{1.3cm}|p{2.2cm}|p{1.6cm}|p{1cm}|}
    \hline
    \textbf{Parameter} & \textbf{Value} & \textbf{Parameter} & \textbf{Value(s)}\\
     \hline
    $h$ & $5$ & $B$ & $2.73$ W\\
    $q$ & $\{0.1,0.5,1,5,10\}$ & $e$ & $1$  \\
    $L_{\text{th}}$ & $80$ & Learning rate & $1e^{-5}$ \\
    $L_{\text{max}}$ & $100$ & Optimizer & AdamW\\
    $A$ & $167$ W & Batch size & $128$ \\
    \hline
    \end{tabular}

    \label{tab:params}
    \end{center}
\end{table}

\begin{itemize}
    \item \(h\): Number of historical traffic samples used as input for prediction. Here, \(h=5\), meaning the model uses the previous five time steps to forecast future traffic.

    \item \(q\): Shape parameter of the Balancing Loss Function (BLF), controlling the asymmetry between underprediction and overprediction penalties. Different values of \(q\) correspond to different operator objectives.

    \item \(L_{th}\): Load threshold used in the cell on-off scheme. If the combined estimated load of a co-located cell pair is below this threshold, one cell can be switched off to save power.

    \item \(L_{max}\): Maximum load that a single active cell can support without causing throughput degradation.

    \item \(A\): Static power consumption component of a cell in the power model, representing load-independent power usage.

    \item \(B\): Dynamic power scaling coefficient in the power model, determining how power consumption increases with traffic load.

    \item \(e\): Spectral efficiency ratio between the co-located cell pair. It scales the transferred load when one cell is switched off.

    \item Learning rate: Step size used by the optimizer during gradient-based training.

    \item Optimizer: Optimization algorithm used for training. AdamW is employed for stable convergence and weight decay regularization.

    \item Batch size: Number of training samples processed simultaneously in one optimization step.
\end{itemize}

\section*{Appendix D}
\subsection*{Balancing Loss Function (BLF) Example}

Here is an illustrative example of how the Balancing Loss Function (BLF) evaluates the difference between the true value ($y$) and the predicted value ($\hat{y}$) for different values of $q$, using

\[
BLF = \max \left\{ \frac{q(y - \hat{y})}{q + 1}, \frac{\hat{y} - y}{q + 1} \right\}.
\]

Assume the true future traffic value is $y = 100$ units. Consider the following prediction scenarios:

\begin{itemize}
    \item \textbf{Underprediction:} $\hat{y} = 90$, resulting in an error of $y - \hat{y} = 10$.
    \item \textbf{Overprediction:} $\hat{y} = 110$, resulting in an error of $\hat{y} - y = 10$.
\end{itemize}

\subsubsection*{Case 1: $q = 1$ (Balanced Penalty)}

\begin{itemize}
    \item For $q = 1$, BLF applies a symmetric penalty to underprediction and overprediction.
    
    \item \textbf{Underprediction Penalty:}
    \[
    \max \left\{ \frac{1 \times 10}{2}, \frac{-10}{2} \right\}
    = \max \{5, -5\} = 5.
    \]

    \item \textbf{Overprediction Penalty:}
    \[
    \max \left\{ \frac{1 \times (-10)}{2}, \frac{10}{2} \right\}
    = \max \{-5, 5\} = 5.
    \]

    \item Thus, underprediction and overprediction incur equal penalties, similar to a balanced regression loss such as MSE.
\end{itemize}

\subsubsection*{Case 2: $q = 0.1$ (Prioritizing Power Savings)}

\begin{itemize}
    \item For $q < 1$, BLF penalizes overprediction more heavily than underprediction.

    \item \textbf{Underprediction Penalty:}
    \[
    \max \left\{ \frac{0.1 \times 10}{1.1}, \frac{-10}{1.1} \right\}
    = \max \{0.91, -9.09\}
    \approx 0.91.
    \]

    \item \textbf{Overprediction Penalty:}
    \[
    \max \left\{ \frac{0.1 \times (-10)}{1.1}, \frac{10}{1.1} \right\}
    = \max \{-0.91, 9.09\}
    \approx 9.09.
    \]

    \item In this case, overprediction is penalized approximately 10 times more strongly than underprediction. This encourages the model to avoid excessively high traffic forecasts, thereby reducing unnecessary cell activations and improving energy efficiency.
\end{itemize}

\subsubsection*{Case 3: $q = 10$ (Prioritizing Service Quality)}

\begin{itemize}
    \item For $q > 1$, BLF penalizes underprediction more heavily than overprediction.

    \item \textbf{Underprediction Penalty:}
    \[
    \max \left\{ \frac{10 \times 10}{11}, \frac{-10}{11} \right\}
    = \max \{9.09, -0.91\}
    \approx 9.09.
    \]

    \item \textbf{Overprediction Penalty:}
    \[
    \max \left\{ \frac{10 \times (-10)}{11}, \frac{10}{11} \right\}
    = \max \{-9.09, 0.91\}
    \approx 0.91.
    \]

    \item In this case, underprediction is penalized approximately 10 times more strongly than overprediction. This encourages the model to avoid forecasting traffic values that are too low, thereby reducing the risk of insufficient resources and network congestion while prioritizing service quality.
\end{itemize}

\end{document}